# VIPriors Object Detection Challenge

Luo Zhipeng[1], Che Lixuan[2]

[1,2]DeepBlue Technology (Shanghai) Co., Ltd

*Abstract*—**This paper is a brief report to our submission to the VIPriors Object Detection Challenge. Object Detection has attracted many researchers' attention for its full application, but it is still a challenging task. In this paper, we study analysis the characteristics of the data, and an effective data enhancement method is proposed. We carefully choose the model which is more suitable for training from scratch. We benefit a lot from using softnms and model fusion skillfully.**

## 1. Introduction

Object Detection has made great process in recent years. Most of the current state-of-art detectors [1, 2] are finetuned from huge amount of annotated data [3]. In many practical application scenarios, due to the limitations of various conditions, we can not get a large number of samples for training . Therefore, it is necessary to study object detection algorithm based on small samples.

VIPriors Object Detection challenge is a object detection challenge, the target of this challenge is optimize for high average precision on the subset of MSCOCO [4] 2017. The challenge of this competition are following:

(1) The number of training dataset is small. There is only 5873 images for training, and 4946 images for validation.

(2) Unbalanced distribution of data categories. There are 80 categories to detect, but the number of each category is extreme imbalance. There are 13085 person in training set, but only 7 hair drier. So is the validation set.

(3) Using any pre-trained model is strictly forbidden. It is difficult for a complex network to converge well from scratch with such a small number of data.

To optimize the average precision, we do a lot on data augmentation, and carefully choose the network structure. Since the commonly used object detection network has been very powerful for feature extraction, we believe that enriching the data set is more effective than modifying the network structure, and our experiments also confirm this. We augment the data set both in pixel level and in spatial level. And wetake categories balance in consideration. For choosing the network ,we do a few experiments, we find that using more shared convolution layers and less fully connection layers is usefully. Some other tips , such as using GN, is useful as well.

## 2. Data Augmentation

We make there type of data augmentation.

1) Categories balance data augmentation. Before the categories balance data augmentation, we count the number of bounding box of each categories in training set. As shown in **Figure** 1, before data balance augmentation, there is an extreme imbalance in the number of different categories. For example, the original train set has 13085 person, but only 7 hair driers, 16 toasters. In order to ensure the diversity of the training data set, we make a maximum of 20 copies of a single image. In order to make the number contrast of each categories more obvious, the category of person, car, bottle, book, cup and chair are not shown in the figure. Before balance data augmentation, there are 13085 person in train data, but only 7 hair drier. The number of hair drier is only 0.5 ‰ of the number of person. After balance data augmentation, there are 161748 person, and 147 hair driers. The number of hair drier is 0.9‰ of the number of person. From **Figure** 1, we can see that the number of other categories has been significantly improved without considering the categories with a large number of categories, such as person, car, chair and so on. And the number difference between categories is significantly reduced. In order to ensure the diversity of images, we augment the copy images in pixel leval. For example, random chang channel, saturation, brightness . The number of different categories is reduced.

We use Mask RCNN [5] as our baseline, with ResNest-50 [6] as backbone and balanced data augmentation as our training data.

2) Pixel data augmentation. Different from only use 3 type of pixel level augmentation (random change channel, saturation, brightness) to copy images in balance augmentation, in pixel data augmentation, we use 30 type of pixel level data augmentation. We randomly generate integers between 0 and 29, each number corresponds to a type of pixel level data augmentation. In **Figure** 2, there are several pixel level data augmentation examples. All pixel level data augmentation are achieve by albumentations [7]. Spatial level data augmentation. We use two types of spatial level data augmentation, shown in **Figure** 3. The first type of spatial level data augmentation is center crop. For center crop, we define the height of output image is h, the width of the output image is w. The height of the original image is defined as image_height, and image_width is the original image's width. Lets x0, y0, x1, y1 denote the coordinate of the top left and bottom right of the original image correspondingly. We use xc, yc donate the coordinate of the center of the original image.We use xmin, ymin, xmax, ymax donate the coordinate of the top left and bottom right of the center crop output image.

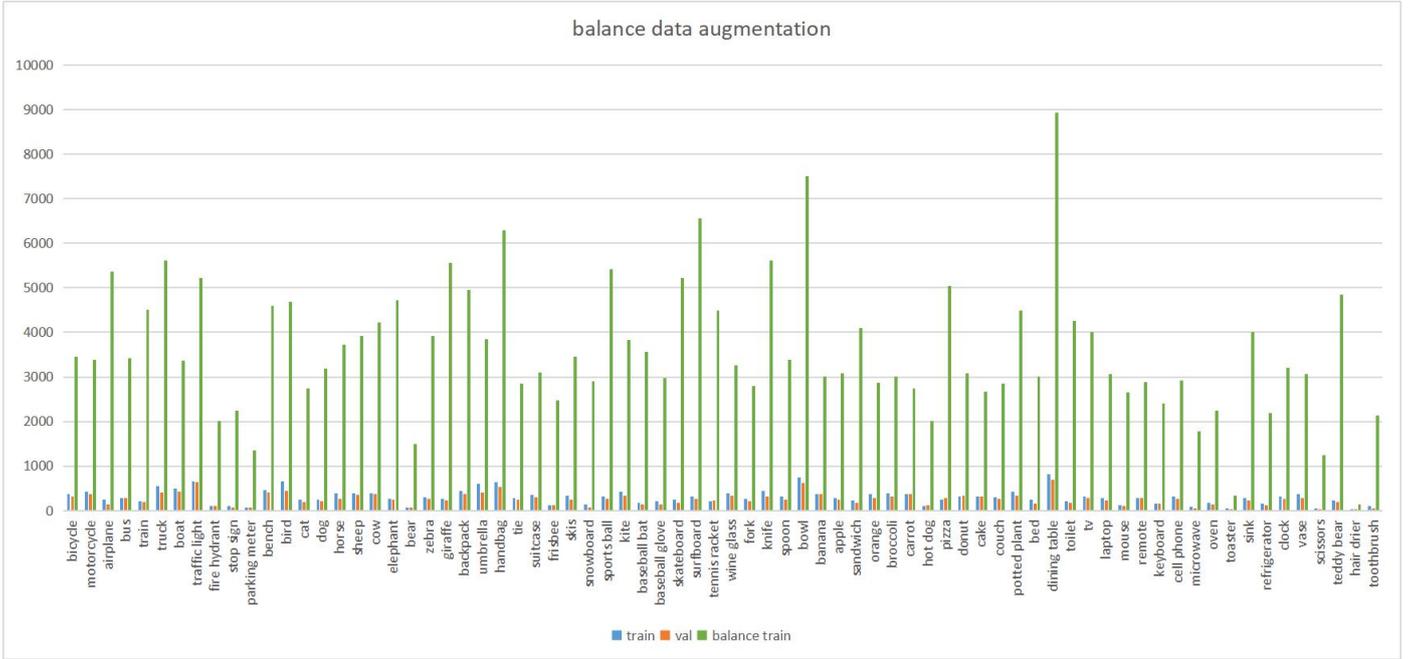

**Figure 1.** Data Distribution of training and validation set. In order to make the number contrast of each categories more obvious, the category of person, car, bottle, book, cup and chair are not shown in the figure.

$$xc = \frac{x0 + x1}{2} = \frac{image\_width}{2}$$

$$yc = \frac{y0 + y1}{2} = \frac{image\_height}{2}$$

$$xmin = xc - \frac{w}{2} = \frac{image\_width - w}{2}$$

$$ymin = yc - \frac{h}{2} = \frac{image\_height - h}{2}$$

$$xmax = xc + \frac{w}{2} = \frac{image\_width + w}{2}$$

$$ymax = yc + \frac{w}{2} = \frac{image\_height + w}{2}$$

The second type of spatial level data augmentation is random sized crop. In this type of augmentation, we input the coordinate of the top left and bottom right, and crop the original image with the input coordinate.In our spatial data augmentation, we use two types of spatial level augmentation twice correspondingly. For center crop, we randomly generate the width and height of the output image, we crop the short side of the original image, and keep the same aspect ratio to get the long side. First the short side of the output image is between 80%-99% of the original image. Second the short side is between 60%-80%. For the type of random sized crop, we control short side of the cropped image as well. We use two different proportion to get two different random sized data augmentation. The key of the spatial level data augmentation is how to deal with the ground truth box at the clipping edge. For these boxes on the clipping edge, we first compute the the IoU between the ground truth and the rest of the box after crop. And then compare the area of the rest box after crop with the area of the output image, we donate the R as the ratio of the box area to output image area. When IoU≥ 0.5 and R > 0.01, we will keep this box during training. When IoU<0.5 and R >0.01, during the training process, we learn the feature of this box, but the loss of this box is not calculated in back propagation. Under the remaining conditions, we will ignore this box.

## 3. Experiment

### A. *Network Selected*

We use Mask R-CNN as baseline network, with ResNet-50 as backbone. But the performance on the validation set is not satisfactory. So we carefully select network structure to make sure the network we selected more suitable for training from scratch. The network we

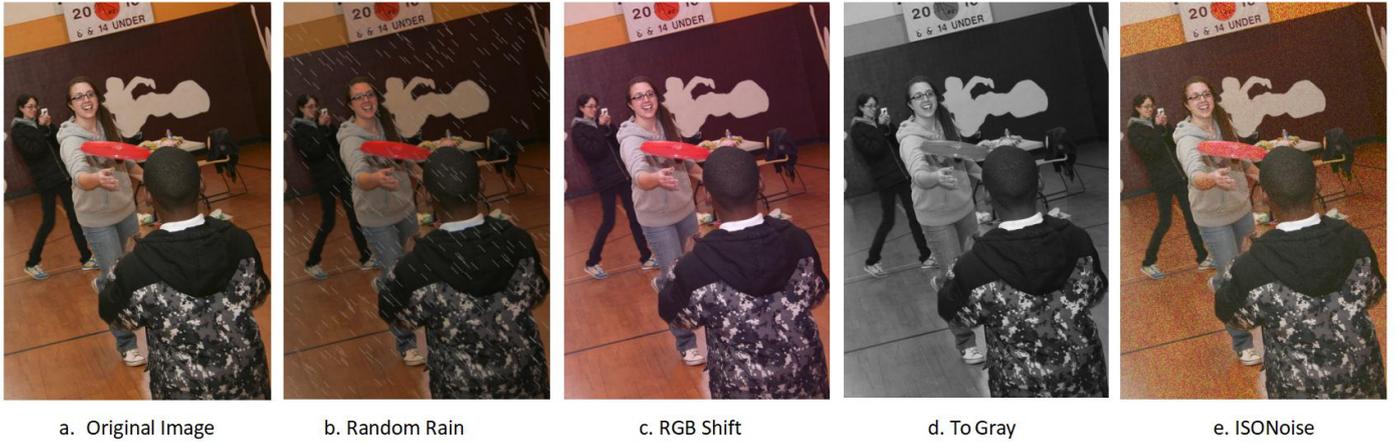

a. Original Image     b. Random Rain     c. RGB Shift     d. To Gray     e. ISONoise

**Figure 2.** Several pixel level data augmentation examples.

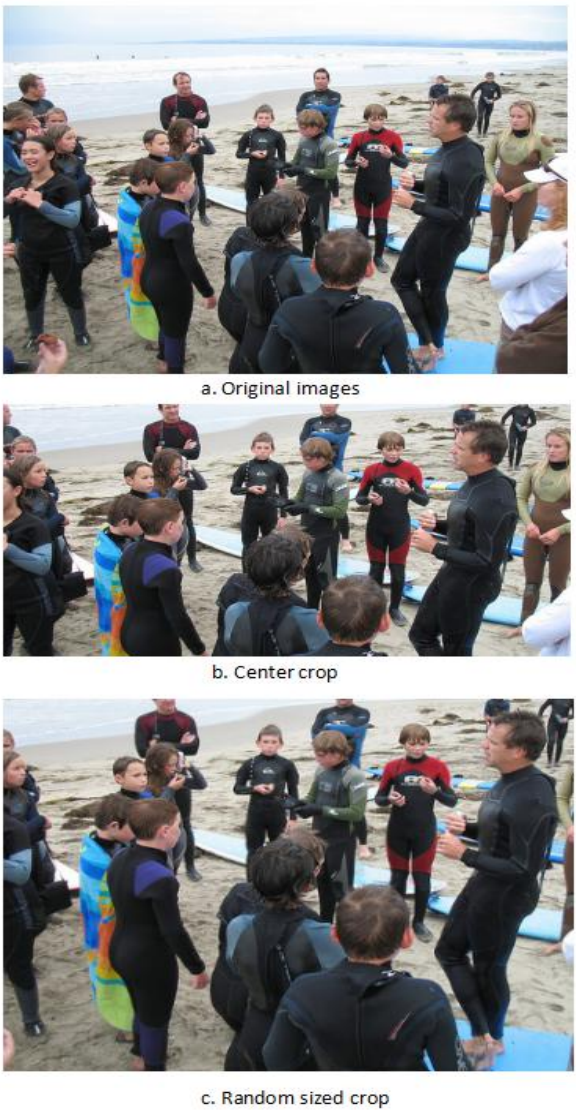

a. Original images

b. Center crop

c. Random sized crop

**Figure 3.** Spatial level data augmentation examples.

select is Scratch Mask RCNN [8]. The differences between Scratch Mask RCNN and original Mask RCNN are as following:

1) Using GN instead of BN. Scratch Mask RCNN use group normalization while Mask RCNN use batch normalization. Group normalization divides channels into groups, and calculates the normalized mean and variance in each group for normalization. The calculation of group normalization has nothing to do with batch, and the accuracy is still stable when the batch size changes greatly. In our experiment, different selection of network models will lead to a big change of batch size. When using a small batch size, the effect of batch normalization will decrease obviously. Therefore, using group normalization is a very effective method for us.

2) Whether to use zero initialization for the last layer of backbone network. Original Mask RCNN use zero initialization for the last layer of backbone network, while the Scratch Mask RCNN not.

3) Different type of bounding box head. The bounding box head of Scratch Mask RCNN contains 4 shared weights convolution layers and 1 shared weights fully connection layer. While the bounding box head of original Mask RCNN contains 2 fully connected layers.

### B. Training strategy and Details

All of our experiments are conducted in 8 GPUs with 12G RAM and 8 GPUs with 32G RAM. According to the previous paper, we know that multi-scale training is a very effective method to improve the results, so we did not use experiments to verify this conclusion again, but directly used multi-scale training. The details of our experiments are as following.

1）Using group normalization. We use Mask RCNN as our baseline, with ResNet-50 as backbone network.

Training the original training data set with multi-scale, We train with image scale (shorter side) randomly sample from [1400, 1800], keeping the aspect ratio of the image. When the network converges sufficiently, we only get 0.083 mAP on the original validation set. With the same parameter settings, using group normalization achieves 0.155 mAP on the validation set, and in the inference process. This experiment verifies the effectiveness of group normalization in improving the results. And in the following experiments, we use group normalization instead of batch normalization.

2）Using more abundant data. We use the balanced categories data augmentation on the training set, copy a single image up to 20 times, and pixel level data augmentation is used for the copied images, such as random change channel, saturation, brightness. The results on the validation set improve from 0.155 to 0.178. Using this checkpoint inference on the test set, and we make our first submit on the test, get 0.174mAP. Through this experiment, we can find that simple image reproduction and color transformation can significantly improve the results.

3）We combine 5873 images of train set and 4946 images of validation set, get total 10819 images data set as original data. We do 30 types of pixel level data augmentation for all 10819 images as pixel augmentation data. Pixel level data augmentation is very limited to improve data diversity, especially training from scratch always requires more iterations. Only pixel level data augmentation may lead the network learn the location of objects during numerous iterations, which will lead to poor generalization performance. Taking this into consideration,we use different parameters do twice center crop and random size crop correspondingly, and a data set with 4 times of the original data is obtained as cropped data. We combine 10819 original data, 10819 pixel augmentation data and 43276 cropped data together, 64914 images in total. In the following experiments, we will all use this data set unless otherwise specified.

4）Using network more suitable for training from scratch. With the same parameter and the same data set (the original 5873 images train set and 4946 images for validation) of the previous experiments, we train Scratch Mask RCNN, ResNet-50 as backbone network. The results on the validation set improve from 0.178 to 0.235. Through this experiment, we verify that Scratch Mask RCNN is effective for improving the results. Due to the limit number of submit on the test set, we did not inference this model on the test set.

After verifying the validity of Scratch Mask RCNN on improving the results, we train Scratch Mask RCNN with our 64914 images data set. With the same parameter setting, we make our second submit on the test set, and get 0.252mAP.

This results is obtained by single scale test, based on past experience multi scale test can improve the results. Due to the limit times of submit on test set, we did not inference the test set with multi scale, but we verify this conclusion on the validation set.

5）Using deeper backbone network. According to the previous paper, we know that using deeper backbone network is another very effective method to improve the results. All previous experiments above use ResNet-50 as backbone network. First, we change the backbone network from ResNet-50 to ResNet-101, we totally iterate the train set 56 epochs, initialize learning rate is 0.02, when iterate to 40 and 50, the learning rate become one tenth of before. During the training process, I want to verify the effect of backbone network, I use the checkpoint of $41^{th}$ epoch, which makes my $4^{th}$ submit on test set. The result of test set inferred with $41^{th}$ epoch is 0.285mAP. And the result of test set inferred with $54^{th}$ epoch is 0.299mAP, which corresponds to my $7^{th}$ sumit on the test set. The results between $4^{th}$ and $7^{th}$ are other experiments.

### C. Other Tips and Experiments

In addition to the strategies mentioned above, we have done some other experiments. Some of these experiments are effective in improving the results, while others are not. In this section, I will introduce these experiments.

1) Adjust the weight of loss function. Take Mask RCNN for example, the total loss of Mask RCNN is defined by the following function.

$$L_{total} = L_{RPN_{cls}} + L_{RPN_{loc}} + L_{box\_head_{cls}} + L_{box\_head_{loc}} + L_{mask}$$

The loss of region proposal network contains classification loss and bounding box regression loss, which are donate by $L_{RPNcls}$ and $L_{RPNloc}$ correspondingly. The output classification loss $L_{box\_head_{cls}}$ and bouding box regression loss $L_{box\_head_{loc}}$ compose the loss of box branch. The mask branch loss donate $L_{mask}$. We can see that there is no coefficient before the amount that makes up the total loss, or all the coefficient is 1. We infer the validate set with our trained model, we find that the location of the predicted bounding box is acceptable, but the predicted categories are not satisfactory. So we adjust the weights of loss function. Based on our trained model, we change the weight of classification loss of box branch from 1 to 1.2, and reduce the weight of mask branch from 1 to 0.3. We finetune the model we trained before with the adjusted loss function. The performance on the validation set improved significantly.

2) Process the predict result with soft-nms. Different from the common nms and soft-nms in deep learning, the

input of our soft-nms is not classified. Due to the poor classification performance, one object always has several predict bounding box with different categories. Process the unclassified results, we can reduce the confidence of some categories of the same object. Our experiments have proved that process the single model's results with unclassified soft-nms, and then fuse the results together is effective on the validation set. Since the model fusion is mentioned, we want to explain that many previous papers have proved the effectiveness of model fusion, so we do not have to prove it by experiment, but we use it directly.

3) Train a model with categories with large mount of ground truth bounding box (for this experiment wee elect 34 categories, including potted plant, bird, horse, sheep, cow, bottle, chair, dining table, bicycle, car, motorcycle, person, umbrella, clock, truck, traffic light, bench, skis, kite, wine glass, cup, knife, bowl, orange, broccoli, carrot, cake, book, vase, backpack, handbag, suitcase, banana and boat). We use the model train by 34 categories as our pre-trained model, and then funtine the model on the balanced category data. But our experiment shows that this method is very limited to improve the results.

4) Use cascade structure in the box head. We experiment cascade structure several times during the process of competition. Train the Scratch Cascade Mask RCNN with the original 5873 images training set will lead to under fitting. The improvement is very limited training the Scratch Cascade Mask RCNN compared with Scratch Mask RCNN. But the results on the 6 times of the original training set is satisfactory.

5) Some other tips, such as multi scale training and test, model fusion, random flip the images when inference, .etc have been proved in many other papers, we will not repect it in this report, we use these tips directly.

## 4. Conclusion

In this competition, we do a lot on data augmentation, and carefully choose the network structure. Skillfully using softnms and nms, and fusing the results of multiple models. Through this competition, we have accumulated the experience of small sample training. Sometimes, rich data is more import than better models.